\documentclass[10pt,twocolumn,letterpaper]{article}

\usepackage{cvpr} 
\usepackage{times}
\usepackage{epsfig}
\usepackage{graphicx}
\usepackage{amsmath}
\usepackage{amssymb}
\usepackage{float}
\usepackage{tablefootnote}
\usepackage{csquotes}
\usepackage{multirow}
\usepackage{comment}
\usepackage{enumitem}
\usepackage{enumerate}
\usepackage{MnSymbol}
\usepackage{float}
\usepackage[table]{xcolor}
\newcommand{\mycc}[1]{\cellcolor{gray!30}#1}
\usepackage{hyperref}
\hypersetup{
    colorlinks=true,
    linkcolor=blue,
    filecolor=magenta,      
    urlcolor=cyan,
    pdftitle={Overleaf Example},
    pdfpagemode=FullScreen,
    }
\usepackage{setspace,lipsum}
\usepackage{booktabs}

\begin{document}

\title{Second Competition on Presentation Attack Detection on ID Card}

\author{ 
Juan E. Tapia\textsuperscript{$\dagger$}\textsuperscript{1},
Mario Nieto\textsuperscript{$\dagger$}\textsuperscript{2}, 
Juan M. Espin\textsuperscript{$\dagger$}\textsuperscript{2}, 
Alvaro S. Rocamora\textsuperscript{$\dagger$}\textsuperscript{2},\\
Javier Barrachina \textsuperscript{$\dagger$}\textsuperscript{2},
Naser Damer\textsuperscript{$\dagger$}\textsuperscript{3,4}, 
Christoph Busch\textsuperscript{$\dagger$}\textsuperscript{1},\\
Marija Ivanovska \textsuperscript{*,5}, 
Leon Todorov \textsuperscript{*,5}, 
Renat Khizbullin\textsuperscript{*,6}, 
Lazar Lazarevich\textsuperscript{*,6},\\ 
Aleksei Grishin\textsuperscript{*,6}, 
Daniel Schulz\textsuperscript{*,7}, 
Sebastian Gonzalez\textsuperscript{*,7}, \\ 
Amir Mohammadi\textsuperscript{*,8}, 
Ketan Kotwal\textsuperscript{*,8}, 
Sebastien Marcel\textsuperscript{*,8},\\ 
Raghavendra Mudgalgundurao \textsuperscript{*,9}, 
Kiran Raja\textsuperscript{*,9}, 
Patrick Schuch\textsuperscript{*9}, \\
Sushrut Patwardhan\textsuperscript{*,9},
Raghavendra Ramachandra\textsuperscript{*,9}, 
Pedro Couto Pereira\textsuperscript{*,10},\\ 
Joao Ribeiro Pinto\textsuperscript{*,10}, 
Mariana Xavier\textsuperscript{*,10}, 
Andrés Valenzuela\textsuperscript{*,11},  
Rodrigo Lara\textsuperscript{*,12},\\ 
Borut Batagelj\textsuperscript{*,13}, 
Marko Peterlin\textsuperscript{*,13}, 
Peter Peer\textsuperscript{*,13}, \\ 
Ajnas Muhammed\textsuperscript{*,14}, 
Diogo Nunes\textsuperscript{*,14},
Nuno Gon{ç}alves\textsuperscript{*,14},\\
\textsuperscript{1}Hochschule Darmstadt (h-da), da/sec-Biometrics and Internet Security Research, Germany,\\
\textsuperscript{2}Facephi company, Spain,\\
\textsuperscript{3}Fraunhofer Institute for Computer Graphics Research IGD, Darmstadt, Germany,\\
\textsuperscript{4}Department of Computer Science, TU Darmstadt, Darmstadt, Germany,\\
\textsuperscript{5}University of Ljubljana, Faculty of Electrical Engineering,\\
\textsuperscript{6}Incode Technologies Inc., San Francisco, CA 94105, USA,\\
\textsuperscript{7}ID VisionCenter (IDVC), Santiago, Chile,
\textsuperscript{8}Idiap Research Institute, Martigny, Switzerland, \\
\textsuperscript{9}Norwegian University of Science and Technology (NTNU), Gj{ø}vik, Norway,\\
\textsuperscript{10}Amadeus, Lisbon, Portugal, \\
\textsuperscript{11}Universidad de Santiago (USACH), Santiago, Chile,\\
\textsuperscript{12}Universidad Andr{é}s Bello, Santiago, Chile,\\
\textsuperscript{13}University of Ljubljana, Faculty of Computer and Information Science,\\
\textsuperscript{14}Institute of Systems and Robotics, Lisbon, Portugal.\\
\textsuperscript{$\dagger$}{\tt\small Organizer}
\textsuperscript{$*$}{\tt\small Competitors}, {\textbf{Full version on IEEE}}
}

\maketitle
\thispagestyle{empty}

\begin{abstract}
This work summarises and reports the results of the second Presentation Attack Detection competition on ID cards. This new version includes new elements compared to the previous one. (1) An automatic evaluation platform was enabled for automatic benchmarking; (2) Two tracks were proposed in order to evaluate algorithms and datasets respectively; and (3) A new ID card dataset was shared with Track 1 teams to serve as the baseline dataset for the training and optimisation. 
The Hochschule Darmstadt, Fraunhofer-IGD, and Facephi company jointly organised this challenge. 
20 teams were registered, and 74 submitted models were evaluated. For Track 1, the \enquote{Dragons} team reached first place with an Average Ranking and Equal Error rate (EER) of ($AV_{Rank}$) of 40.48\% and 11.44\% EER, respectively. For the more challenging approach in Track 2, the \enquote{Incode} team reached the best results with an $AV_{Rank}$ of 14.76\% and 6.36\% EER, improving on the results of the first edition of 74.30\% and 21.87\% EER, respectively. These results suggest that PAD on ID cards is improving, but it is still a challenging problem related to the number of images, especially of bona fide images.
\end{abstract}

\section{Introduction}
PAD-ID Card 2025 is the second competition in this series, offering an independent assessment of the current state-of-the-art algorithms that are (re-)trained on an open dataset and benchmarked on real operational, sequestered datasets. This information and results will be fully available after the competition is closed, which means that researchers can continue to improve their solutions and compare the results with the winner's solution and baselines.

PAD-ID Card 2025 proposed two different types of participation and evaluation, organised in two tracks:

\subsection{Track 1: Synthetic Dataset for Training}

Track 1 of this challenge allows participants to train or retrain their models on a shared synthetic dataset, made publicly available at the beginning of the competition. This dataset was specifically designed to simulate real-world ID verification scenarios, while maintaining full control over data generation and annotation. This track's primary goal is to identify the best algorithm based on a common dataset.

\subsection{Track 2: Open-set dataset for Training}
Track 2 of this challenge allows participants to train their models on any open set dataset available on the state of the art and complement their data with private datasets. The main goal of this track is to identify how the number of bona fide images, countries, and diversity of attacks are relevant to obtain generalisation capabilities. This track opens a challenge for commercial applications. 

\section{Datasets}
\label{sec:datasets}

\subsection{Synthetic ID Card Generation Process}

The dataset was generated using digitally generated ID card templates that mimic a wide range of identity documents from several countries. We initially collected frontal-facing ID templates and manually cleared all personal identifiable information, such as names, dates, faces, and signatures, using Adobe Photoshop\footnote{\url{https://www.adobe.com/es/products/photoshop.html}}. This ensured a clean and artefact-free basis for further manipulations.

Next, synthetic identities were generated using a combination of computer vision techniques and public datasets. Seamless cloning methods~\cite{poisson} were used to insert new data, including randomly selected faces, synthetic signatures, names, and alphanumeric characters, into the cleared templates. These assets were sourced from a combination of public face datasets~\cite{face-dataset-1, face-dataset-2, face-dataset-3} and signature datasets~\cite{signature-dataset-1, signature-dataset-2, signature-dataset-3}, allowing the creation of unique and realistic ID cards (see examples in Figure~\ref{fig:attacks-maded}).

\subsection{Shared Synthetic Dataset}

The dataset provided for Track 1 includes 12,000 images divided between various classes representing both bona fide and attack scenarios. Only the front side of the ID cards is used. The composition is as follows:

\begin{itemize}
    \item \textbf{3,000 bona fide images}, created by printing ID card designs with synthetic face images on PVC cards and subsequently capture.
    \item \textbf{3,000 screen attack images}, captured by displaying the printed and recaptured PVC ID cards on a variety of digital screens (e.g. laptops, tablets, smartphones) and photographing them again in various resolutions.
    \item \textbf{1,000 gray copy images}, consisting of gray-scale printed reproductions on glossy and standard paper with different thicknesses, and photographing them again in various resolutions.
    \item \textbf{2,000 colour copy images}, consisting of colour printed reproductions on glossy and standard paper with different thicknesses and photographing them again in various resolutions.
    \item \textbf{1,500 physical composite attacks}, generated by physically altering the user's face in the PVC ID cards with paper cutouts of different faces, pasted in both rectangular and irregular shapes and photographing them again in various resolutions.
    \item \textbf{1,500 digital composite attacks}, created by digitally altering the synthetic ID cards, replacing facial regions with other synthetic identities using seamless cloning.
\end{itemize}

Each image class in the shared dataset is uniformly distributed across a large set of synthetic identities, covering 24 countries and 155 unique card templates. The full distribution of image types and associated user identities for Track 1 is detailed in Table~\ref{tab:db-track1-dist}.

\begin{table}[h]
\scriptsize
\centering
\caption{Distribution Track 1 Database}
\label{tab:db-track1-dist}
\begin{tabular}{|l|c|c|}
\hline
          & Nº Imgs & Nº Unique Subjects \\ \hline
Bona fide &   3,000    &  155     \\ \hline
Screen attack &   3,000    &  155      \\ \hline
Gray Print     &   1,000    &  155     \\ \hline
Colour Print    &   2,000    &  155     \\ \hline
Physical Composite     &   1,500   &  155     \\ \hline
Digital Composite     &   1,500   &  155     \\ \hline
Total     &   12,000   &  155     \\ \hline
\end{tabular}%
\end{table}

\begin{figure}[h]
\centering
\subfloat[ESP ID Card Screen Display]{
  \includegraphics[width=35mm,height=24mm]{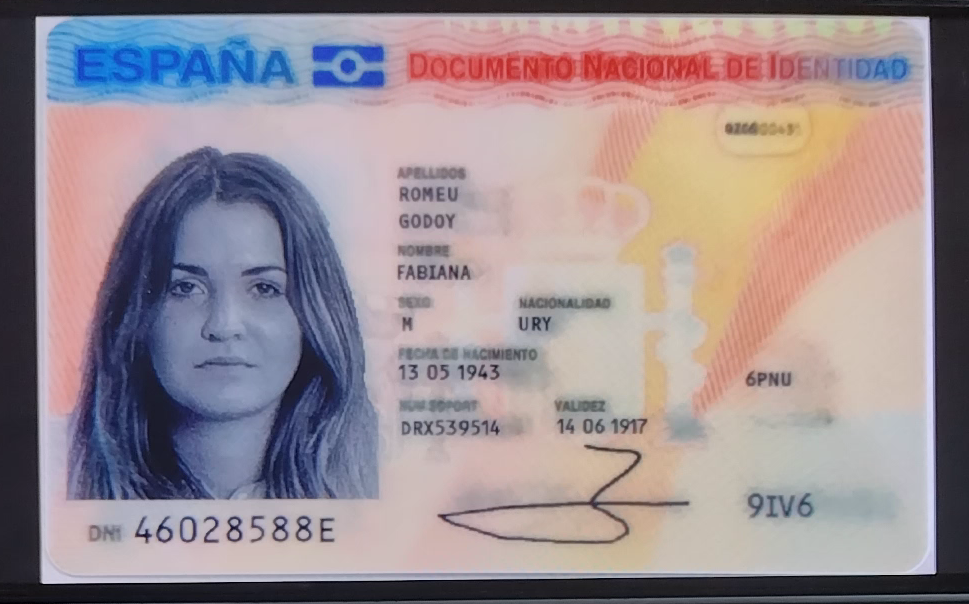}\label{fig:spoof-esp}
}
\subfloat[ESP ID Card Gray Print]{
  \includegraphics[width=35mm,height=24mm]{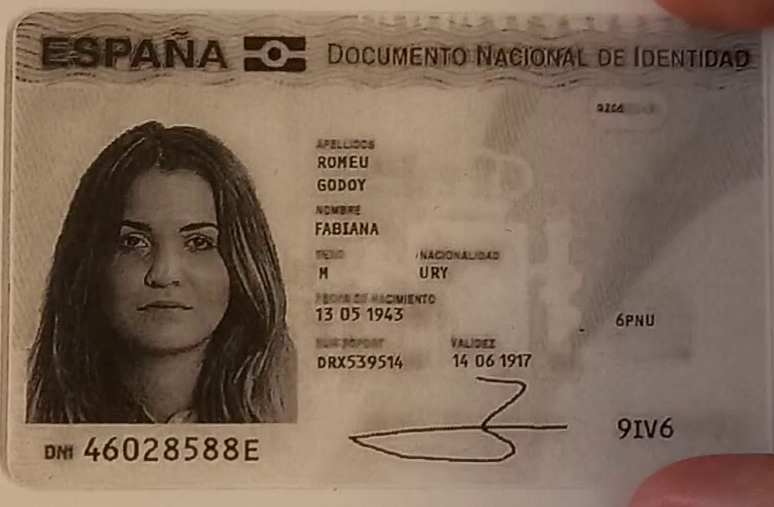}\label{fig:copy-esp}
}

\subfloat[ESP ID Card Colour Print]{
  \includegraphics[width=35mm,height=24mm]{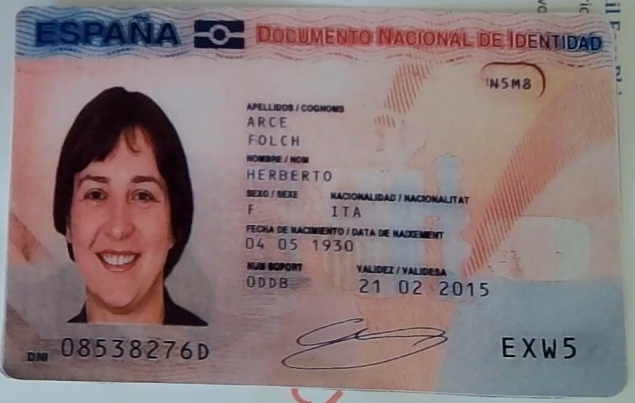}\label{fig:color-copy-esp}
}
\subfloat[ESP ID Card Physical Manipulation]{
  \includegraphics[width=35mm,height=23mm]{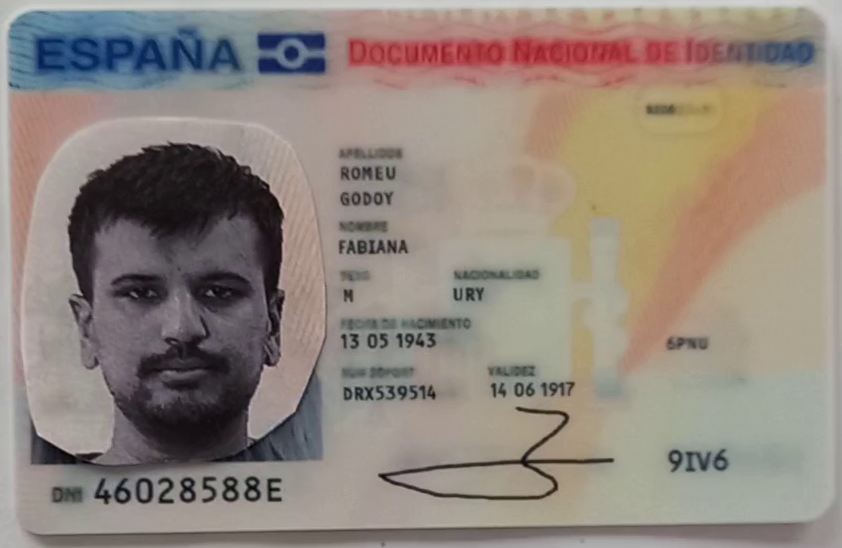}\label{fig:composite-manual-esp}
}

\subfloat[ESP ID Card Digital Manipulation]{
  \includegraphics[width=35mm,height=23mm]{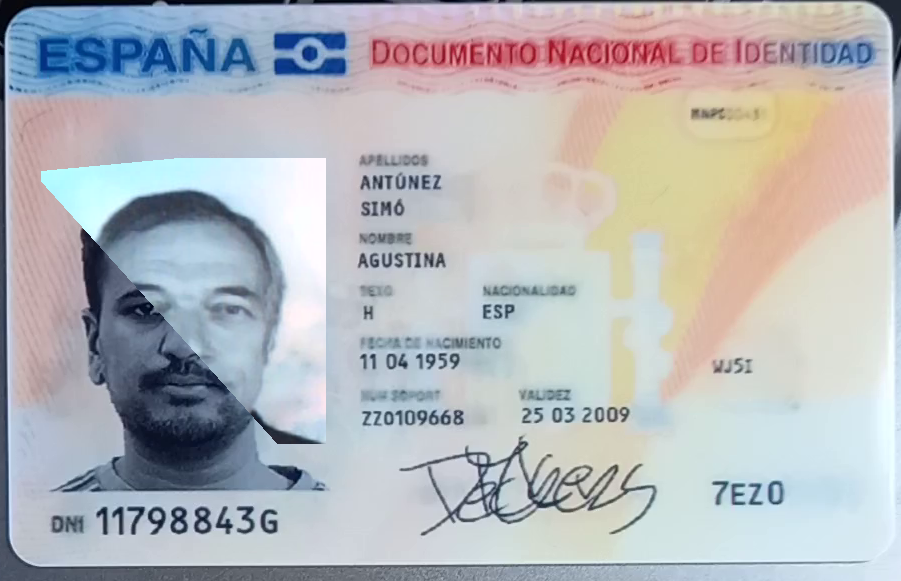}\label{fig:composite-digital-esp}
}
\subfloat[ESP ID Card Digital Bona fide]{
  \includegraphics[width=35mm,height=23mm]{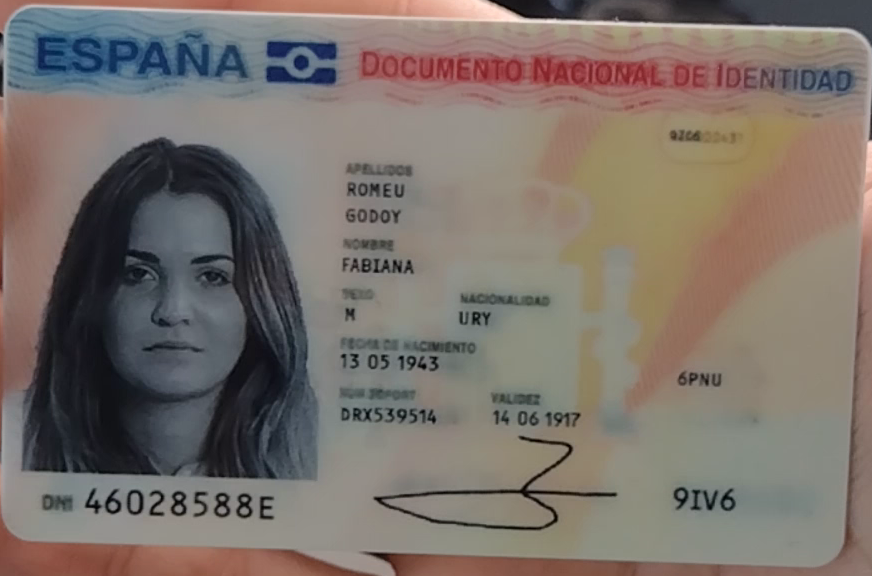}\label{fig:bona-esp}
}
\caption{Examples of the different presentation and manipulation attacks generated for the Track 1 Database.}
\label{fig:attacks-maded}
\end{figure}

This carefully annotated synthetic dataset forms the foundation for participants to develop presentation attack detection models in Track 1, with a controlled and reproducible setup that reflects realistic deployment challenges.
Examples of the attacks generated for the dataset can be seen in Figure \ref{fig:attacks-maded}.

\subsection{Test Dataset Description}

The evaluation of all submitted systems, including the baseline systems, was conducted using a sequestered test set that remains unchanged from the previous edition of the competition~\cite{challenge-past}. This decision ensures consistency across evaluations and allows for fair comparisons with the prior year's results.

The test set comprises a total of 23,851 images representing ID cards from \textbf{4 different countries}, Chile (CHL), Guatemala (GTM), Panama (PAN), and Mexico (MEX). For each country, the dataset includes bona fide samples as well as multiple types of presentation attacks. Specifically, the attacks are categorised into three main types: \textit{screen} attacks (captured from various digital displays), \textit{print attacks} (both standard paper prints and PVC), and \textit{composite attacks} (digital manipulations and forgeries).

Each image has undergone a detailed quality curation process to reflect the diverse acquisition conditions. This includes variations in lighting, image resolution, compression artefacts, and the presence or absence of ICAO compliance. Attack samples were generated using both manual and automated procedures to simulate realistic adversarial scenarios. Printed attacks were created from both glossy paper and PVC cards, while screen attacks were recorded using different devices and screen types, including smartphones, tablets, and laptops, across various resolutions.

Furthermore, the dataset contains contributions from 5,000 unique user IDs, with images spread across 20 distinct ID versions. This variety ensures that models are tested against a wide range of visual and semantic variability.

Importantly, no injection or digital-only attacks were considered in this evaluation. The focus remains on the physical presentation of forged identity documents, aligning with the goals of real-world deployment scenarios in online document authentication systems.

Tables~\ref{tab:track1_baseline},~\ref{tab:db-baseline-exp1},~\ref{tab:db-baseline-exp2}, and \ref{tab:db-baseline-exp3} describe the details of each group of images used for the different baselines, which are explained in Section~\ref{sec:baseline}.

\section{Submission Process}
\label{sec:submission_process}
To handle submissions, we hosted a custom evaluation platform based on EvalAI~\cite{EvalAI}. This platform handles the challenge participants and submissions, sending evaluation requests to our private server. This platform allowed us to evaluate multiple submissions per participant. We evaluated 25 submissions on Track 1 with 9 participants and 49 on Track 2 with 7 participants.

An API embedded into a Docker image was used to evaluate submissions. This API has an endpoint that receives an image and returns a continuous score in the range [0, 1] where 0 represents total confidence that the image is a presentation attack while 1 represents total confidence that the image is a bona fide. We assume that all images can be processed, i.e., there are no bad-quality or corrupted images. Any processing error is treated with a score of 0, i.e., an attack is detected. The evaluation proceeds by making requests to the API with every image in the dataset. Once all images are processed, the metrics presented in section~\ref{sec:eval} are calculated and sent back to the platform to update the leaderboard. Only the best submission per participant is shown there and reported in this paper.

\section{Performance Evaluation Criteria}
\label{sec:eval}

The detection performance of biometric PAD algorithms is standardised by ISO/IEC 30107-3\footnote{\url{https://www.iso.org/standard/79520.html}}. The most relevant metrics for this study are Attack Presentation Classification Error Rate (APCER), Bona fide Presentation Classification Error Rate (BPCER), and BPCER\textsubscript{AP}. Those metrics determine the error rates when classifying an instance between bona fide and the different Presentation Attack Instrument Species (PAIS).  

The APCER metric measures the percentage of attack presentations incorrectly classified as bona fide for each different PAIS. The computation method is detailed in Equation~\ref{eq:apcer}, where the value of $N_{PAIS}$ corresponds to the number of attack presentation samples, $RES_{i}$ is $1$ if the $i$th image is classified as an attack, or $0$ if it was classified as a bona fide presentation according to a predefined threshold.

\begin{equation}\label{eq:apcer}
    {APCER_{PAIS}}=1 - \frac{1}{N_{PAIS}}\sum_{i=1}^{N_{PAIS}}RES_{i}
\end{equation}

On the other hand, the BPCER metric measures the proportion of bona fide presentations wrongly classified as attacks. The BPCER can be computed using Equation~\ref{eq:bpcer}, where $N_{BF}$ is the number of bona fide presentation samples, and $RES_{i}$ takes the same values described earlier for the APCER metric. The two metrics determine the system's performance and are subject to a specific operation point. 

\begin{equation}\label{eq:bpcer}
    BPCER=\frac{\sum_{i=1}^{N_{BF}}RES_{i}}{N_{BF}}
\end{equation} 

Finally, BPCER\textsubscript{AP} and the Equal Error Rate (EER) are used to analyse the PAD system's performance for a specific operating point. The latter is the operating point where APCER and BPCER are equal. This operating point corresponds to the intersection with the diagonal line in a Detection Error Trade-off (DET) curve, which is also reported for all the experiments. On the other hand, the BPCER\textsubscript{AP} is the BPCER value when the APCER is $100/AP$. In this work, we use: BPCER\textsubscript{10}, BPCER\textsubscript{20} and BPCER\textsubscript{100}, which correspond to APCER values of 10\%, 5\% and 1\%, respectively.

An average ranking $AV_{Rank}$ determines the winning team. A weighting factor is selected for each BPCER@APCER to increase the metric's contribution in the most challenging operational points based on the threshold. Specifically, BPCER\textsubscript{10} was weighted by 0.2, BPCER\textsubscript{20} by 0.3, and BPCER\textsubscript{100} by 0.5. The team with the lowest $AV_{Rank}$ won the competition. This metric weighted the BPCER\textsubscript{10,20,100} as follows to emphasise high security applications:
\vspace{-0.3cm}

\begin{equation}\label{eq:avrank}
\scriptstyle
    AV_{Rank}=BPCER_{10}\times0.2+ BPCER_{20}\times0.3 + BPCER_{100}\times0.5
\end{equation} 

\section{Baseline Description}
\label{sec:baseline}
This section describes the architectures and training process of the baseline models and participant submissions for both Track 1 and Track 2.
According to the available datasets, different baselines were defined for each track to explore the real conditions and possible scenarios of today's state-of-the-art~\cite {GONZALEZ-PR, markham2023openset}.

\subsection{Baseline for Track 1}
For Track 1, a single baseline configuration was established using only the official dataset shared by the competition organisers. The baseline was designed to reflect a controlled yet realistic training scenario by adhering strictly to the provided data.

The training and validation splits were built by partitioning the shared Track 1 database based on unique ID cards per country. Specifically, 80\% of the ID cards from each country were used for training, and the remaining 20\% constitute the validation set. This ensured that identities in the validation set were not seen during training, providing a more accurate assessment of generalisation performance. Table \ref{tab:track1_baseline} provides a summary of the dataset used for the Track 1 Baseline. 

\begin{table}[H]
\scriptsize
\centering
\caption{Baseline dataset - Training/Validation/Test using shared dataset}
\label{tab:track1_baseline}
\begin{tabular}{|c|c|c|c|c|}
\hline
          & Train & Val & Test & Total \\ \hline
Bona fide &   2,397    &  603  & 8,000 &   11,000    \\ \hline
Composite &   2,451    &  549  & 4,993 &   7,993   \\ \hline
Print     &   2,407    &  593  & 5,969 &   8,969    \\ \hline
Screen    &   2,351    &  649  & 4,889 &   7,889    \\ \hline
Total     &   9,606   & 2,394   & 23,851 &   35,851   \\ \hline
\end{tabular}%
\end{table}

The model architecture selected for this baseline was \textit{EfficientNetV2-S}~\cite{efficientnetv2}, initialised with ImageNet~\cite{deng2009imagenet} pre-trained weights.

Unlike the previous challenge, where the best results were obtained using the~\textit{MobileViTv2} \cite{howard2017mobilenets} architecture, for this track we opted for a convolutional architecture over a vision transformer due to the significantly smaller amount of training data available in Track 1. Vision transformers, such as MobileViT, generally require large-scale datasets to fully exploit their capacity and achieve good generalisation. In contrast, EfficientNetV2-S is a lightweight and highly efficient convolutional neural network that performs well in low-data regimes.

The classification task was defined as a four-class problem:~\textit{bona fide} (PVC ID cards), ~\textit{Screen} display attacks,~\textit{Prints} (Copy and ColourCopy), and~\textit{Composite}. Training was conducted with class balancing to ensure that each batch contained an equal number of bona fide and attack samples, mitigating any potential bias due to class imbalance.

All input images were preprocessed by cropping the ID document region with an additional 5\% padding to preserve the background information of the captured image. The resulting images were resized to $384 \times 384$ pixels to match the input requirements of the chosen architecture.

To improve model robustness and simulate real-world capture conditions, a series of data augmentations were applied during training with a 50\% probability. These augmentations included horizontal flipping, random brightness and contrast adjustments, JPEG compression artefacts, Gaussian filtering, and random hue shifts.

This baseline configuration represents a standard supervised training setting using only the Track 1 shared dataset, serving as a reference point for evaluating more complex or generalised approaches in this competition.

\subsection{Baseline for Track 2}

According to the challenge that represented the previous edition, we have chosen to reuse the same baseline configurations presented in our prior work~\cite{challenge-past}. This decision was further motivated by the low performance of the participating models in last year's challenge, where baseline models significantly outperformed all the submitted solutions.

Three baselines were defined based on different combinations of private and open-set datasets, aiming to simulate a variety of real-world scenarios and assess model generalizability:

\begin{itemize}
    \item \textbf{Baseline 1}: Training is performed exclusively on a private dataset that includes both bona fide and attack samples. The model evaluation is conducted using a sequestered test dataset comprising unseen samples from multiple countries and attack types.    
    \item \textbf{Baseline 2}: The training set includes only bona fide images from private datasets, combined with attack samples from open-set sources such as MIDV-500 and MIDV-Holo. This configuration simulates a situation where bona fide samples are well-curated internally, while attack examples are drawn from publicly available datasets.   
    \item \textbf{Baseline 3}: A more mixed training configuration that uses both private datasets and open-set sources (MIDV-500 and MIDV-Holo) for both bona fide and attack classes. This setup is intended to test the benefits of greater dataset diversity.
\end{itemize}

The preprocessing pipeline follows the procedure proposed in~\cite{challenge-past} and begins with ID card segmentation to ensure a standardised input format. Each image is then resized to $384 \times 384$ pixels. A series of augmentations are applied to simulate various capture conditions, including horizontal flipping, brightness and contrast variation, JPEG compression, Gaussian blur, hue shift, and grayscale conversion. These augmentations are designed to increase model robustness and reduce overfitting.
\vspace{-0.1cm}

All models were trained with three output classes:~\textit{bona fide},~\textit{composite}, and \textit{print plus PVC and screen} grouped as a unified class to simplify the classification task. The MobileViTv2~\cite{howard2017mobilenets} architecture, using ImageNet~\cite{deng2009imagenet} weights, yielded the strongest performance in our previous experiments~\cite{challenge-past} and is used as the primary reference model in subsequent evaluations.

All baselines are evaluated using the sequestered test set composed of 23,851 images described in Section \ref{sec:datasets}, representing a diverse set of ID cards from four countries and four different Presentation Attack Instrument Species (PAIS). These baselines provide a strong and standardised benchmark for Track 2 participants.


Table \ref{tab:db-baseline-exp1} summarises the dataset used for Baseline 1, which considers only private datasets for training.

\begin{table}[H]
\scriptsize
\centering
\caption{Baseline 1 dataset - Training/Validation using a private dataset}
\label{tab:db-baseline-exp1}
\begin{tabular}{|c|c|c|c|c|}
\hline
          & Train & Val & Test & Total \\ \hline
Bona fide &   64,933    &  11,458  &   5,000  &  81,391     \\ \hline
Composite &   90,101    &  15.900  &   4,993  &  110,994     \\ \hline
Print     &   30,398    &  5,364   &   8,969  &  44,731     \\ \hline
Screen    &   82,232    &  14,511  &   4,889  &  101,632     \\ \hline
Total     &   267,664   &  47,233  &   23,851 &  338,748     \\ \hline
\end{tabular}%
\end{table}

Table \ref{tab:db-baseline-exp2} summarises the dataset used for Baseline 2, which considers only open-set datasets for training, plus private bona fide presentations. This experiment considers MIDV-500 and MIDV-Holo.

\begin{table}[H]
\scriptsize
\centering
\caption{Baseline 2 dataset - Training/Validation on bona fide presentations from a private dataset and attacks from public datasets.}
\label{tab:db-baseline-exp2}
\begin{tabular}{|c|c|c|c|c|}
\hline
          & Train & Val & Test & Total \\ \hline
Bona fide &  64,933   &  11,458 &  5,000  &  81,391     \\ \hline
Composite &  9,283    &  1,638  &  4,993  &  15,914     \\ \hline
Print     &  26,623   &  4,698  &  8,969  &  40,290     \\ \hline
Screen    &  19,608   &  3,460  &  4,889  &  27,957     \\ \hline
Total     &  120,447  &  21,254  & 23,851 &  165,552     \\ \hline
\end{tabular}%
\end{table}

Finally, Table \ref{tab:db-baseline-exp3} provides a summary of the dataset used for Baseline 3, which considers private datasets (267,664 images) plus open-set datasets for training (55,514) images. This experiment finds MIDV-500 and MIDV-Holo.

\begin{table}[H]
\scriptsize
\centering
\caption{Baseline 3 dataset - Training/Validation on a mixture of private and public datasets}
\label{tab:db-baseline-exp3}
\begin{tabular}{|c|c|c|c|c|}
\hline
          & Train & Val & Test & Total \\ \hline
Bona fide &   64,933   &  11,458   &   5,000  &  81,391     \\ \hline
Composite &   99,384   &  17,538   &   4,993  &  121,915     \\ \hline
Print     &   57,021   &  10,062   &   8,969  &  76,052     \\ \hline
Screen    &   101,840  &  17,971   &   4,889  &  124,700     \\ \hline
Total     &   323,178  &  57,029   &   23,851 &  404,058     \\ \hline
\end{tabular}%
\end{table}

\begin{table*}[h]
\scriptsize
\centering
\caption{Track 1 summary challenge results. All the results are in \%.}
\label{tab:track1-results}
\begin{tabular}{@{}lllllll@{}}
\toprule
Rank & Team & EER & BPCER\textsubscript{10} & BPCER\textsubscript{20} & BPCER\textsubscript{100} & AVRank \\ \midrule
1 & \mycc{\textbf{dragons}} & \mycc{\textbf{11.34}} & \mycc{\textbf{13.21}} & \mycc{\textbf{24.39}} & \mycc{\textbf{61.04}} & \mycc{\textbf{40.48}} \\
2 & Idiap & 14.12 & 18.30 & 30.81 & 61.43 & 43.62 \\
3 & Baseline & 16.51 & 27.80 & 45.26 & 77.11 & 57.70 \\
4 & IDCH & 21.66 & 39.36 & 51.66 & 71.75 & 59.25 \\
5 & Asmodeus & 24.01 & 43.08 & 57.28 & 80.73 & 66.16 \\
6 & UNLJ-FRI-FE & 26.53 & 51.29 & 65.13 & 83.95 & 71.77 \\
7 & IDVC-PAD-IDCARD & 22.41 & 46.86 & 65.59 & 91.33 & 74.71 \\
8 & VISTeam & 36.30 & 61.88 & 73.09 & 89.09 & 78.85 \\
9 & PADINO-v2 & 32.97 & 72.05 & 84.23 & 97.04 & 88.20 \\ \bottomrule
\end{tabular}
\end{table*}

\begin{figure*}
\centering
        \begin{subfigure}[b]{0.26\textwidth}
                \centering                \includegraphics[width=.96\linewidth]{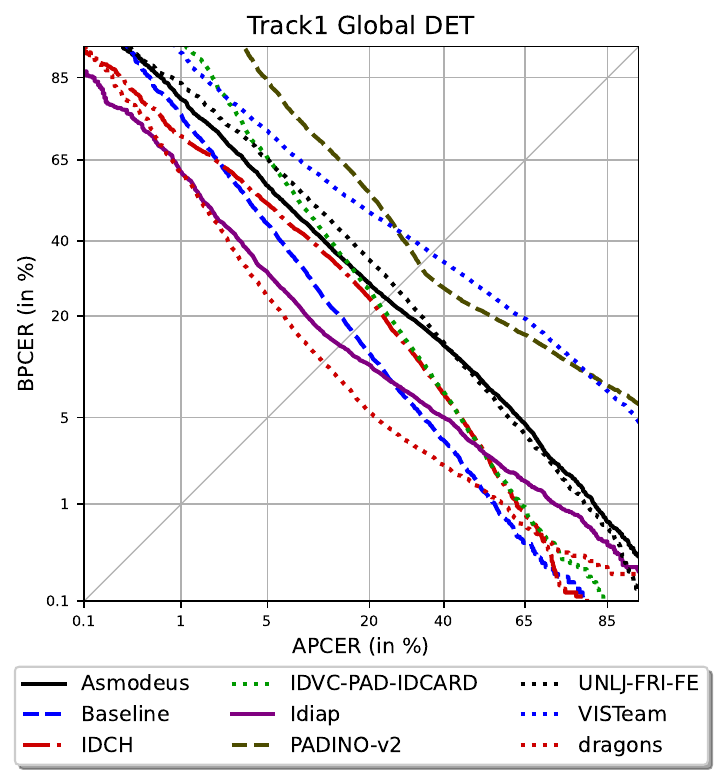}\label{fig:global_track1}
                \caption{Global}
        \end{subfigure}%
        \begin{subfigure}[b]{0.26\textwidth}
                \centering                \includegraphics[width=.96\linewidth]{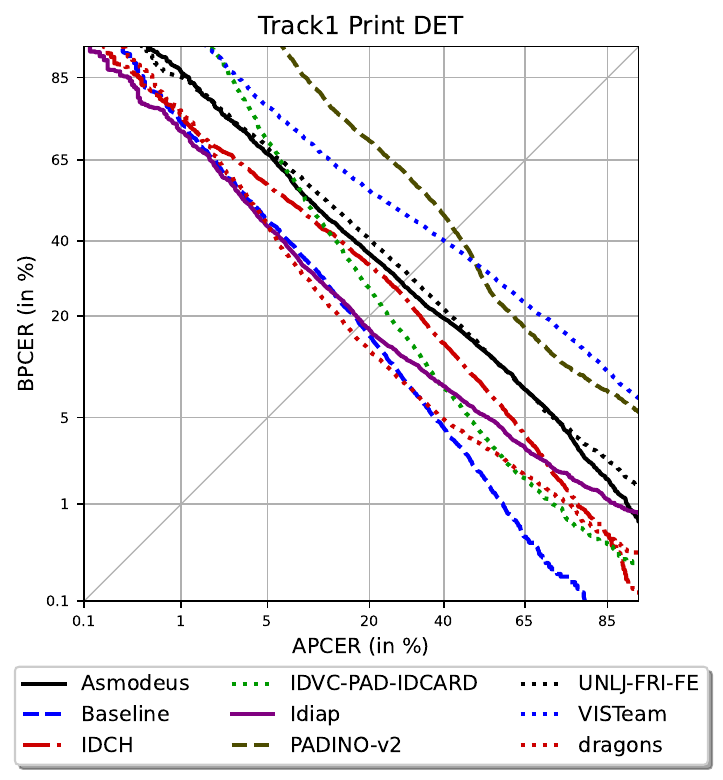}
                \caption{Print}
                \label{fig:print_track1}
        \end{subfigure}%
        \begin{subfigure}[b]{0.26\textwidth}
                \centering                \includegraphics[width=.96\linewidth]{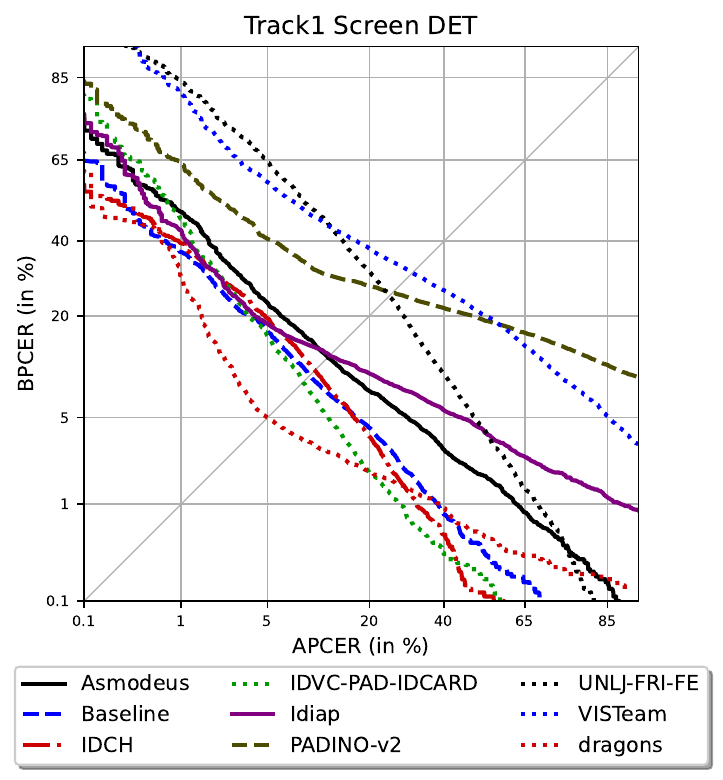}
                \caption{Screen}
                \label{fig:screen_track1}
        \end{subfigure}%
        \begin{subfigure}[b]{0.26\textwidth}
                \centering                \includegraphics[width=.96\linewidth]{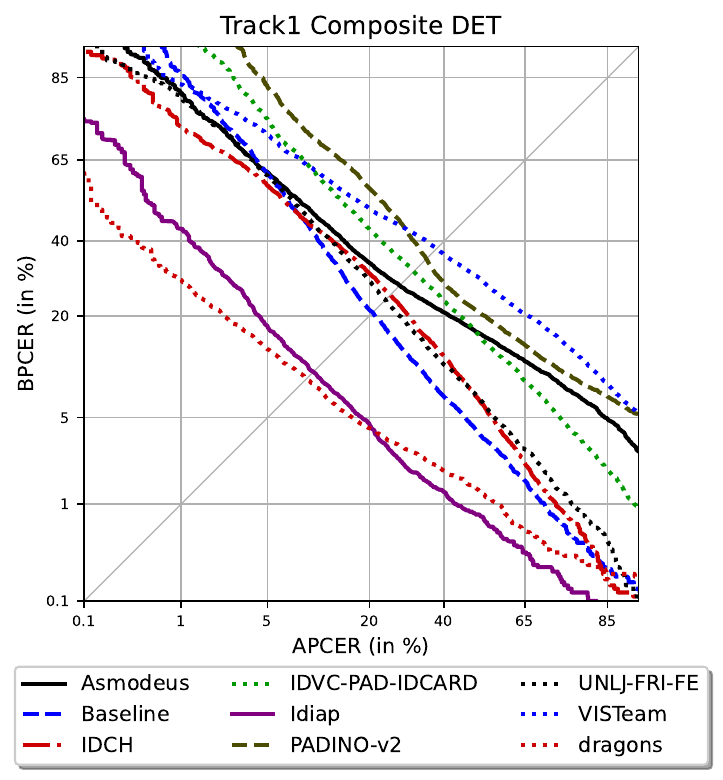}
                \caption{Composite}
                \label{fig:composite_track1}
        \end{subfigure}
        \begin{subfigure}[b]{0.26\textwidth}
                \centering                \includegraphics[width=.96\linewidth]{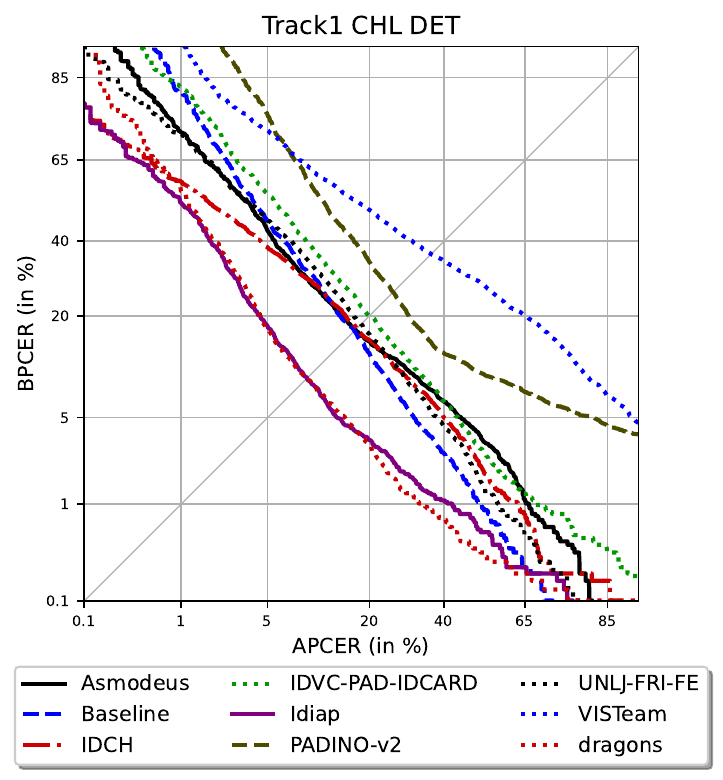}
                \caption{CHL}
                \label{fig:chl_track1}
        \end{subfigure}%
        \begin{subfigure}[b]{0.26\textwidth}
                \centering                \includegraphics[width=.96\linewidth]{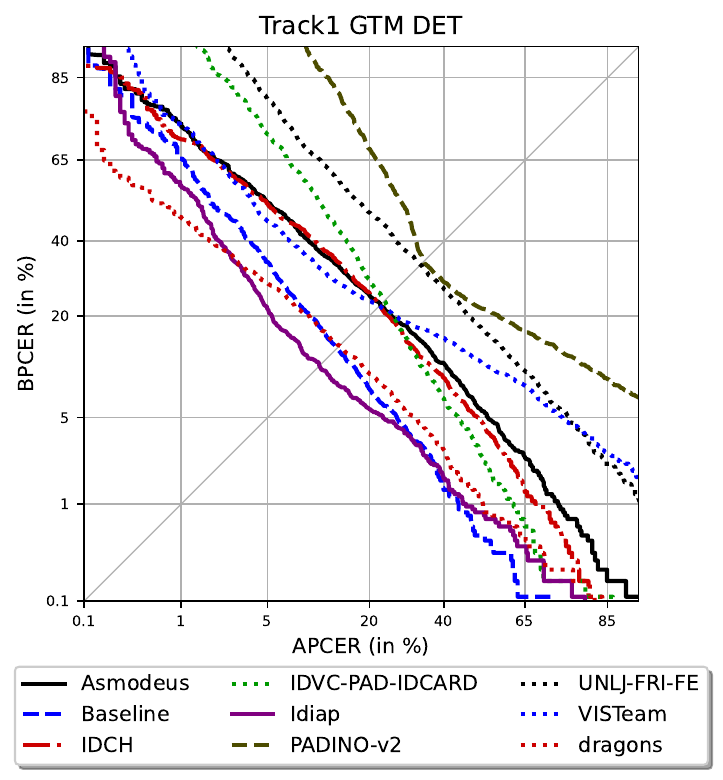}
                \caption{GTM}
                \label{fig:gtm_track1}
        \end{subfigure}%
        \begin{subfigure}[b]{0.26\textwidth}
                \centering                \includegraphics[width=.96\linewidth]{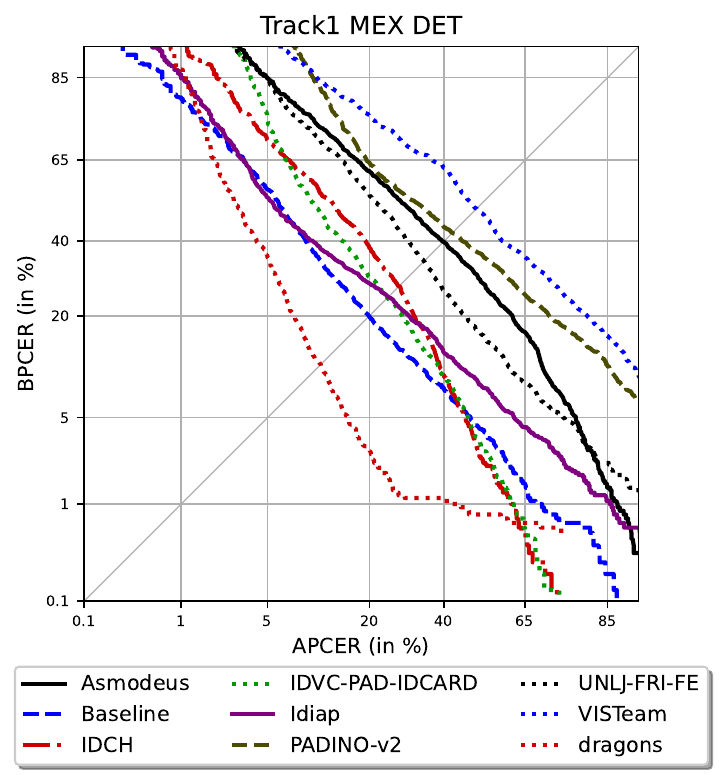}
                \caption{MEX}
                \label{fig:mex_track1}
        \end{subfigure}%
        \begin{subfigure}[b]{0.26\textwidth}
                \centering                \includegraphics[width=.96\linewidth]{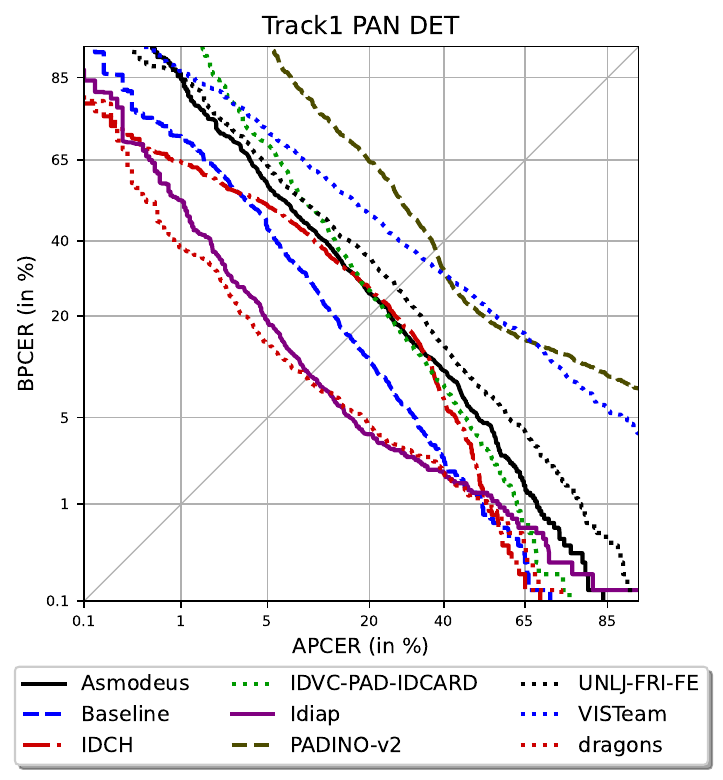}
                \caption{PAN}
                \label{fig:pan_track1}
        \end{subfigure}
        \caption{Track 1 global, per PAIS (b),(c) and, (d) and per country DET curve results (e),(f), and (g). }\label{fig:track1}
\end{figure*}

\begin{table*}[h]
\centering
\scriptsize
\caption{Track 2 summary challenge results. All the results are in \%.}
\label{tab:track2-results}
\begin{tabular}{@{}lllllll@{}}
\toprule
Rank & Team & EER & BPCER\textsubscript{10} & BPCER\textsubscript{20} & BPCER\textsubscript{100} & AVRank \\ \midrule
1 & \mycc{\textbf{Incode}} & \mycc{\textbf{6.36}} & \mycc{\textbf{2.56}} & \mycc{\textbf{9.08}} & \mycc{\textbf{23.04}} & \mycc{\textbf{14.76}} \\
2 & Baseline-1 & 6.07 & 3.06 & 7.90 & 23.64 & 14.80 \\
3 & Baseline-2 & 7.10 & 5.10 & 9.76 & 22.68 & 15.29 \\
4 & Baseline-3 & 8.86 & 7.98 & 12.88 & 27.64 & 19.28 \\
5 & IDVC-PAD-IDCARD & 23.87 & 52.30 & 63.74 & 76.44 & 67.80 \\
6 & \mycc{Best-PAD-2024} & 21.87 & 46.06 & 65.82 & 90.70 & \mycc{74.30} \\ 
7 & Idiap & 31.94 & 57.22 & 70.20 & 87.72 & 76.36 \\
8 & InvestigAI & 34.64 & 70.32 & 81.98 & 94.26 & 85.79 \\
9 & PADINO-v2 & 45.64 & 83.50 & 100 & 100 & 96.70 \\ 

\bottomrule
\end{tabular}
\end{table*}

\begin{figure*}
\centering
        \begin{subfigure}[b]{0.26\textwidth}
                \centering
                \includegraphics[width=.96\linewidth]{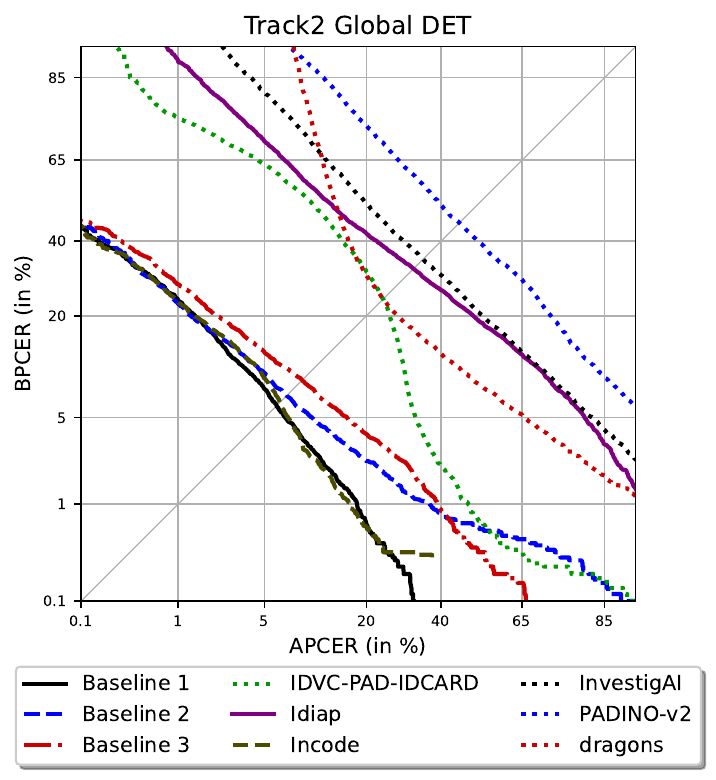}\label{fig:global_track2}
                \caption{Global}
        \end{subfigure}%
        \begin{subfigure}[b]{0.26\textwidth}
                \centering
                \includegraphics[width=.96\linewidth]{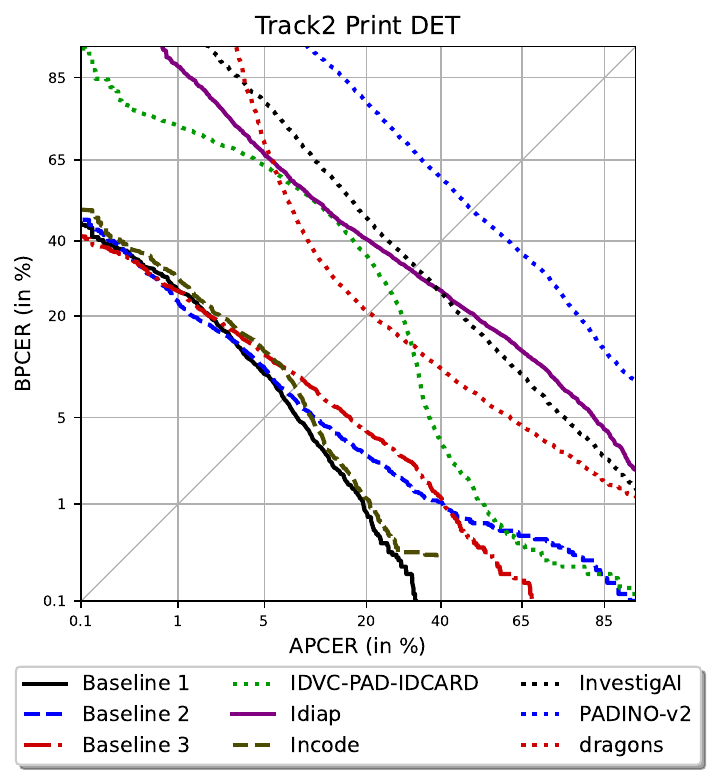}
                \caption{Print}
                \label{fig:print_track2}
        \end{subfigure}%
        \begin{subfigure}[b]{0.26\textwidth}
                \centering
                \includegraphics[width=.96\linewidth]{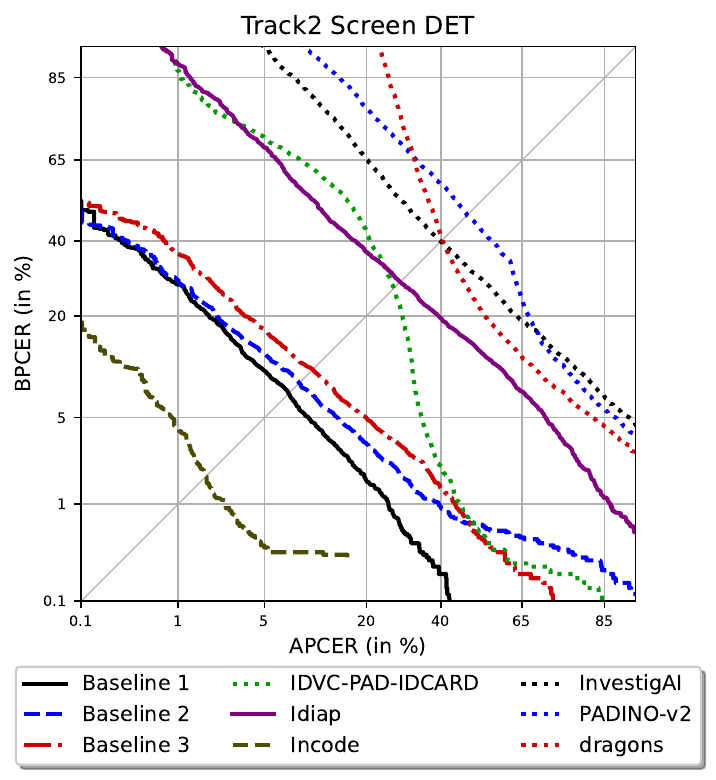}
                \caption{Screen}
                \label{fig:screen_track2}
        \end{subfigure}%
        \begin{subfigure}[b]{0.26\textwidth}
                \centering               \includegraphics[width=.96\linewidth]{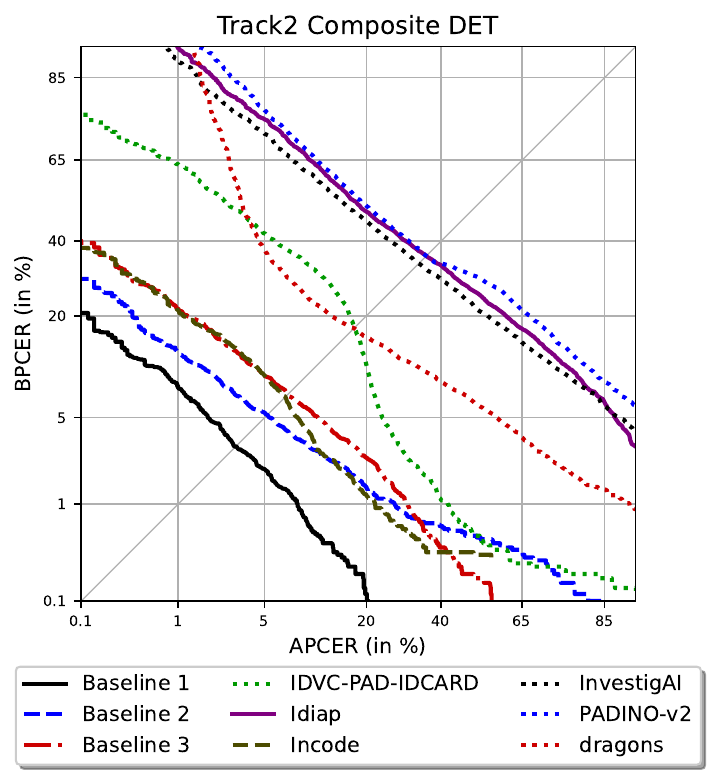}
                \caption{Composite}
                \label{fig:composite_track2}
        \end{subfigure}
        \begin{subfigure}[b]{0.26\textwidth}
                \centering                \includegraphics[width=.96\linewidth]{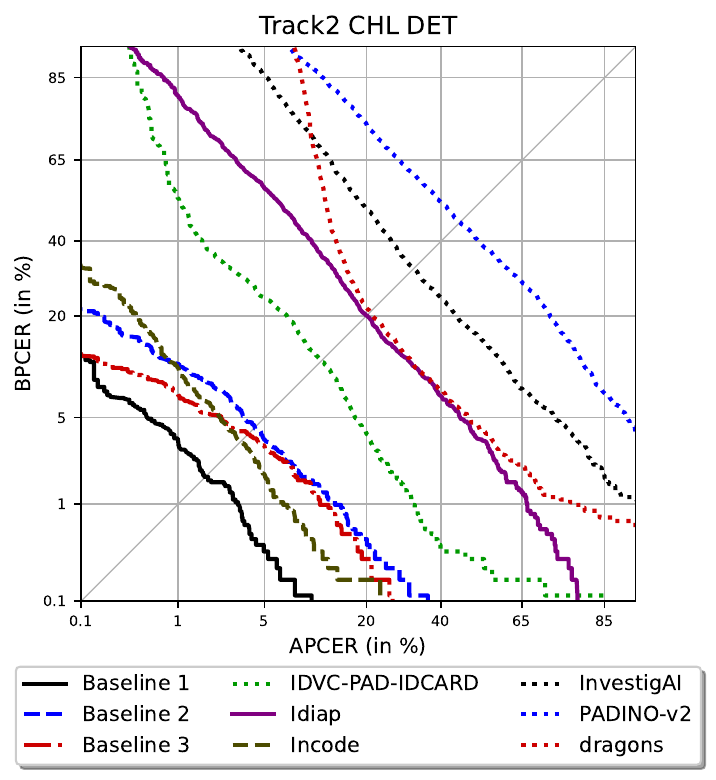}
                \caption{CHL}
                \label{fig:chl_track2}
        \end{subfigure}%
        \begin{subfigure}[b]{0.26\textwidth}
                \centering                \includegraphics[width=.96\linewidth]{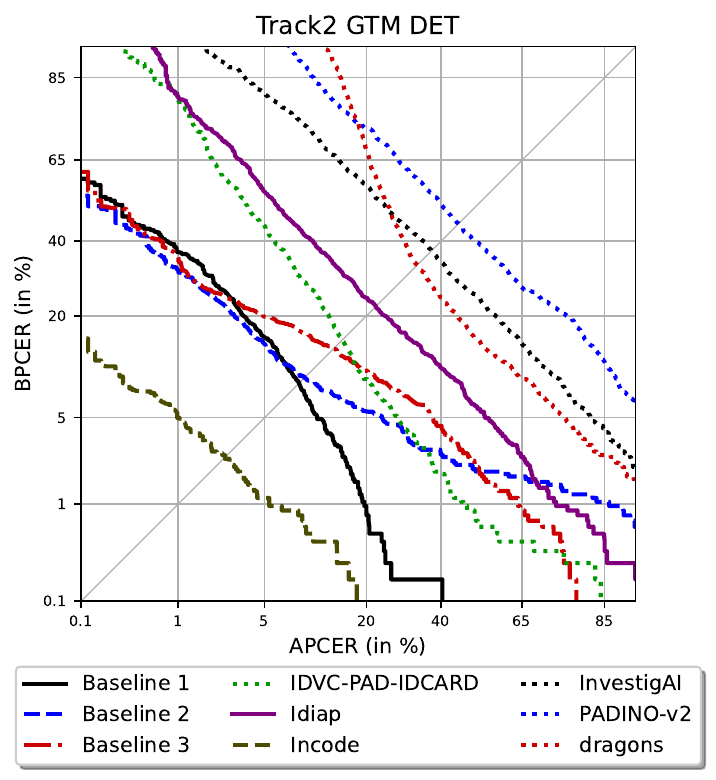}
                \caption{GTM}
                \label{fig:gtm_track2}
        \end{subfigure}%
        \begin{subfigure}[b]{0.26\textwidth}
                \centering                \includegraphics[width=.96\linewidth]{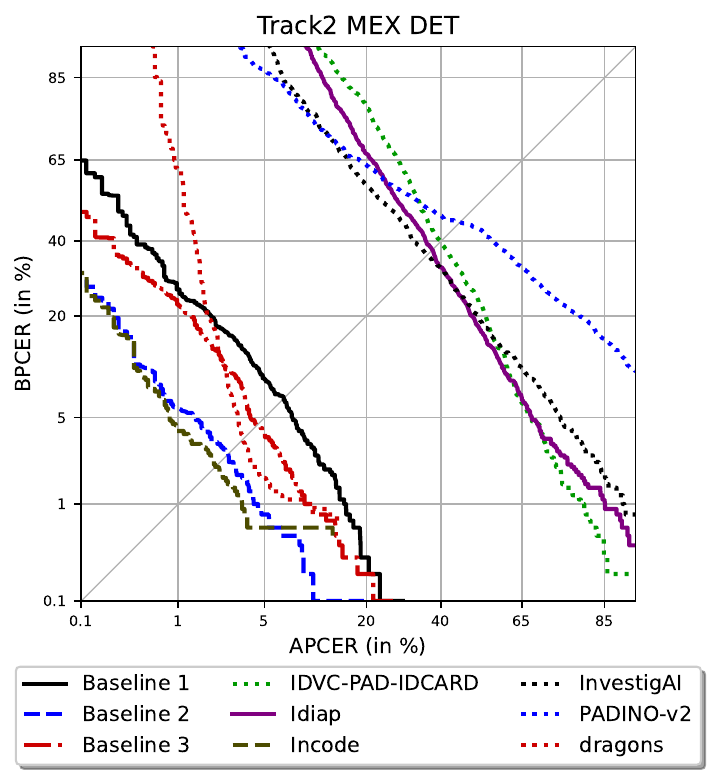}
                \caption{MEX}
                \label{fig:mex_track2}
        \end{subfigure}%
        \begin{subfigure}[b]{0.26\textwidth}
                \centering                \includegraphics[width=.96\linewidth]{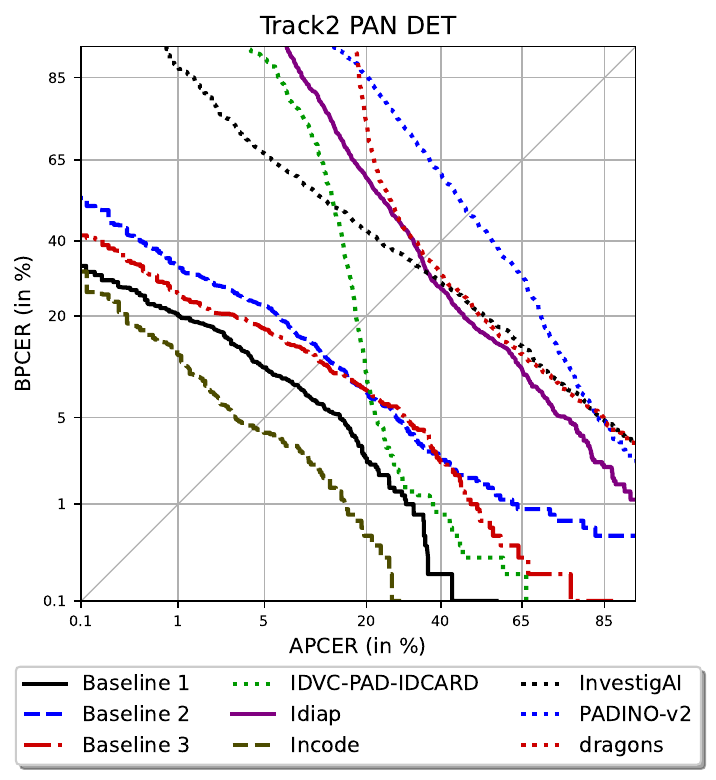}
                \caption{PAN}
                \label{fig:pan_track2}
        \end{subfigure}
        \caption{Track 2 global, per PAIS (b),(c) and, (d) and per country DET curve results (e),(f) and, (g).}\label{fig:track2}
\end{figure*}
\vspace{-0.3cm}

\subsection{Submission to the Competition}
For this competition, 20 teams were registered, but only 10 presented valid models with competitive results. Each team is described as follows:
\subsubsection{\textit{Dragons}}
This team for Track 1 used the provided training dataset containing uncropped ID cards on random backgrounds. We first fine-tune YOLOv8-seg to detect and segment ID cards in a pixel-wise manner~\cite{yolo}. For this purpose, they used 500 randomly selected and manually annotated samples from the IJCB-IDCard-PAD2025 training dataset. During inference, it was leveraging the segmentation mask provided by the YOLO segment to find contours and estimate a $4$ point polygon. Using this polygon, they crop out the ID card and warp it into a square of size $336\times336$ pixels. With this data preprocessing pipeline, the dragons team adapt a pre-trained CLIP visual encoder~\cite{cherti2023reproducible} using LORA~\cite{Lora}. They used a ViT-L/14 backbone for feature extraction with a custom-defined, fully connected binary classification head on top. The network was optimised with a weighted cross-entropy loss for $5$ epochs.

\subsubsection{\textit{Incode}}
For Track 2, Incode's ID-Card-PAD model is an ensemble of 2 CNN classifiers and 1 transformer-based classifier, each trained for a specific presentation attack type: screen, paper or document alteration, in a document-type-guided fashion. The fraud detection models are accompanied by a pre-processing stage for document detection, cropping, and document type recognition. The fraud detection models were trained with our proprietary dataset, which consists mainly of real-world legit and fraudulent images of identity documents, and extended with synthetically generated screen, paper, and alteration samples to balance class distribution towards fraud. The training dataset distribution for our model is 130K images of authentic documents, 37K screen spoofs, 35K paper spoofs, and 16K modified documents.
\vspace{-0.3cm}

\subsubsection{\textit{IDVC}}

For Track 1, team IDVC-PAD-IDCARD employed the DinoV2 ViT-B/14 pre-trained model as a feature extractor ~\cite{dinov2}, adding a linear layer at the end for the classification task. As required, the images were resized to $266\times266$ pixels and normalised using the mean and standard deviation values derived from the ImageNet dataset. For data augmentation, random horizontal flips with a probability of 50\%, and random 90°, 180°, and 270° rotations with a probability of 20\% were applied.

For Track 2, the IDVC-PAD-IDCARD method is composed of two stages. The first stage is an agnostic ID card and passport detector based on YOLOv3, trained on the MIDV-2020 dataset and a private dataset containing Chilean ID cards. The second stage, the PAD detection method, uses the cropped ID cards from the first stage as input. This method is based on the SwinTransformer algorithm, a hierarchical vision transformer that processes images efficiently by using window-based self-attention mechanisms and shifted windows to improve scalability and performance. The method was trained on a private dataset with approximately 140K images, distributed among bona fide, manual, and automatic composites, prints, and screens. Most of the data in this dataset is from Chilean ID cards, with a small amount (about 250 images per country, including bona fide and attack samples) from three other countries. The input for the SwinTransformer is a $224\times224$ image, which has various data augmentations applied during the training process, such as 90-degree rotations, brightness and contrast adjustments, translations, and more.
\vspace{-0.3cm}

\subsubsection{\textit{Idiap}}

The Idiap team\footnote{This work was funded by InnoSuisse 106.729 IP-ICT project.} submitted a Clip (clip-vit-large-patch14) model with linear probing fine-tuned with LoRA. All attention and fully connected matrices of the CLIP model were fine-tuned using LoRA. Card images were detected and cropped  using a commercial model from PXL Vision. Card images were 0-padded to make them a square and resized to $224\times224$ images. 

For Track 2, the same CLIP model was trained on both the competition dataset and an internal PAD dataset. This internal dataset consisted of real ID cards and passports (mainly from Switzerland and its neighbors), and attacks were print and screen attacks. There were around 48K bona fide images, 4K print attacks, and 2K screen attacks for training.


\subsubsection{\textit{IDCH}}
The IDHC team proposes a dual-supervision PAD approach that exploits global image classification and per-pixel examination to improve presentation attack detection accuracy. The method allows the model to identify both overall image authenticity (boan fide) and attacks based on localised regions containing artefacts.
The architecture is EfficientNetV2-S, a pre-trained convolutional neural network on ImageNet. It has the best balance between computation and feature representation ability.

\subsubsection{\textit{UNLJ-FRIFE}}
This team proposed a solution uses a multi-branch deep learning architecture combining ConvNeXt-Base backbone with specialised forensic analysis branches: (a)Semantic Branch: ConvNeXt-Base (ImageNet-22k pretrained) for high-level feature extraction, (b) Texture Branch: CNN analysing fine-grained texture patterns and printing artefacts, and (c) Edge Branch: Dedicated analysis of edge inconsistencies and compression artefacts. The final pipeline: RetinaFace/OpenCV face detection → portrait cropping → $384\times384$ normalisation → multi-class prediction (bona fide/composite/print/screen) → PAD score derivation.
\vspace{-0.3cm}

\subsubsection{\textit{VISTeam}}
The VISTeam has not disclosed a description of the model.
\vspace{-0.3cm}

\subsubsection{\textit{PADINOV2}}
The PADINO-v2 team proposed for Track 1, a system designed for robust detection of ID card presentation attacks using a multi-stage deep learning pipeline: (a) Feature Extraction and Segmentation: A DINOv2 Vision Transformer, fine-tuned with LoRA, processes the input image to extract features, producing a global representation (CLS token) and a segmentation map that highlights the ID card region. (b)Mask-Based Cropping: The segmentation map creates a binary mask, allowing the system to crop and resize the original image to focus on the ID card, reducing background clutter. (c) Classification: The cropped image is input into a lightweight MobileViTv2 classifier, and its output is concatenated with the DINOv2-CLS-token. This combined feature is processed through a multi-layer perceptron (MLP) to normalise the features, enabling either bona fide or attack classification.

For Track 2, the Segment-Border-Source (SBS) system is an effective pipeline for detecting ID card presentation attacks. It operates as follows: (a)Segmentation: A MobileViT-v3 network identifies the ID card region in the input image, creating a binary mask, (b)Mask-Based Cropping, the mask crops the image, removing the background and focusing on the ID card. And (c) two-stage classification: A MobileViT-v3 border classifier first assesses the processed image to detect composite attacks. If none are found, a second MobileViT-v3 source classifier determines if the image is bona fide or an attack. 
This multi-stage approach employs MobileViT-v3 models for segmentation and classification, ensuring efficient and accurate ID card attack detection. Each model was trained separately before being integrated into the complete framework.
\vspace{-0.3cm}

\subsubsection{\textit{Asmodeus}} 
For Track 1, the Asmodeus team fine-tuned the MobileViTv2 architecture to detect presentation attacks. During training, they enhanced the model’s robustness by synthetically generating screen-based attack variations from bona fide samples. This was achieved by simulating moiré patterns and print-attack-like artefacts using a combination of colour jitter, geometric distortions, and diverse image augmentations. These augmentations mimic real-world artefacts encountered in screen replays and print attacks, enabling the model to better discriminate between bona fide and attack inputs.

\subsubsection{\textit{InvestigAI}}
For Track 2, the InvestigAI team used a pre-trained YOLOv8-World model with a single class label, "Document", to detect and crop regions containing ID cards from input images. These cropped regions were then processed using an EfficientNetV2-Small model, pre-trained on ImageNet and fine-tuned on ID card and passport images from MIDV-Holo, DLC2021, and IDNet (explicitly excluding driver's licenses and morphing attack samples). For training, The InvestigAI team used 136,032 images (32,419 bona fide and 103,613 presentation attacks), with 62,498 images allocated for validation (14,754 bona fide and 47.744 presentation attacks). It applies dropout-based regularisation and several data augmentation techniques (colour jittering, Gaussian blur, random perspective transformation, and horizontal flipping) to improve model generalisation.

\section{Results}

\textbf{Competition Winners:} The PAD-ID Card 2025 has two winners, corresponding to each track. For Track 1, the \enquote{Dragons} team wins, with an $AV_{Rank}$ of 40.48\% using a CLIP model. For Track 2, the \enquote{Incode} team wins with an $AV_{Rank}$ of 14.76\% using an ensemble of classifiers based on VisionTransformer. All the results are summarised in Tables \ref{tab:track1-results} and \ref{tab:track2-results}.

\subsection{Track 1}
For Track 1, according to our test set, the most challenging attack is the print PAIS, while the most difficult ID card country is Panama. This is to be expected since the Track 1 training dataset does not include any samples from Panama and shows lower generalisation capabilities.

Figure~\ref{fig:track1} illustrates the DET curves for Track 1. Global DET illustrates all teams as a binary classifier of bona fide versus attacks. Figure~\ref{fig:track1}(b), (c), and (d) illustrate the results of each team considering bona fide and each attack separately, reported from print, screen, and composite attacks, respectively. Also, figure~\ref{fig:track1} illustrates the DET curves for each country, considering Chile, Mexico, Guatemala, and Panama.

\subsection{Track 2}
For Track 2, according to our test set, the most challenging attack is the screen ID cards, and Panama's ID card is the most difficult per country. The Incode team was the only one team that surpassed the baseline on Track 2 and improved on the results from the PAD-IDCard-2024 competition. All the remaining team's results still have room for improvement. It is relevant to highlight that the IDVC-PAD-IDCARD team also surpasses the best results from the 2024 edition, but not for the 2025 edition.

Figure~\ref{fig:track2} illustrates the DET curves for Track 2. Global DET illustrates all teams as a binary classifier of bona fide versus attacks. Figure \ref{fig:track2}(b), (c), and (d) illustrate the results of each team considering bona fide and each attack separately, reported from print, screen, and composite attacks, respectively. Also, figure~\ref{fig:track2} illustrates the DET curves for each country, considering Chile, Mexico, Guatemala, and Panama.

We can also highlight the generalisation capabilities shown by foundation models in Track 1, such as CLIP. The best-performing models of Track 1 are both fine-tuned CLIP, showing good generalisation to both unseen countries and real bona fides. However, in the more challenging Track 2, the best-performing model is an ensemble of CNN models and Vision Transformers trained with a large number of bona fide samples. This shows that combining global features from CNN with localised features from patches derived from Vision Transformer helps to generalise challenging scenarios and countries. This is reinforced by the substantial dataset used by foundation models.  Having access to large datasets for training is still the main issue for PAD ID cards algorithms in generalising to real cases.

\section{Conclusion}
The number of ID cards, the number of subjects, and the number of images are relevant to improving the results. The ID card images used as bona fide make a significant difference, which opens the challenge and debate about obtaining and increasing the number of images and countries while complying with privacy protection restrictions. 

The focus of this edition was to highlight the algorithms and dataset sizes. However, the efficiency of the model must also be considered in future editions, because many foundation models can obtain very good results, but at the cost of an expensive consumption of resources, taking in some cases 1 or 2 minutes of inference time per image. This time is a restriction for any online digital onboarding process.
In summary, the competition was a success from the organisers' point of view because it increased the number of participants and triplicate the number of model evaluations compared to the first edition, showing the increasing interest in this topic.  

\section*{Acknowledgements}
This competition was supported by Hochschule Darmstadt (h-da), Facephi, R\&D area, and Fraunhofer-IGD. 
Further, this work was supported by the European Union’s Horizon 2020 research and innovation programs under grants 101121280 (EINSTEIN) and CarMen (101168325), and the German Federal Ministry of Education and Research and the Hessian Ministry of Higher Education, Research, Science and the Arts within their joint support of the National Research Center for Applied Cybersecurity ATHENE.

{\small
\bibliographystyle{ieee}
\bibliography{egbib}

\begin{thebibliography}{10}\itemsep=-1pt

\bibitem{signature-dataset-1}
V.~L. Blankers, C.~E. v.~d. Heuvel, K.~Y. Franke, and L.~G. Vuurpijl.
\newblock {ICDAR} signature verification competition.
\newblock In {\em 10th International Conference on Document Analysis and Recognition}, pages 1403--1407, 2009.

\bibitem{face-dataset-1}
U.~Cheema and S.~Moon.
\newblock Sejong face database: A multi-modal disguise face database.
\newblock {\em Computer Vision and Image Understanding}, 208-209:103218, 2021.

\bibitem{cherti2023reproducible}
M.~Cherti, R.~Beaumont, R.~Wightman, M.~Wortsman, G.~Ilharco, C.~Gordon, C.~Schuhmann, L.~Schmidt, and J.~Jitsev.
\newblock Reproducible scaling laws for contrastive language-image learning.
\newblock In {\em Proceedings of the IEEE/CVF Conference on Computer Vision and Pattern Recognition}, pages 2818--2829, 2023.

\bibitem{deng2009imagenet}
J.~Deng, W.~Dong, R.~Socher, L.-J. Li, K.~Li, and L.~Fei-Fei.
\newblock Imagenet: A large-scale hierarchical image database.
\newblock In {\em IEEE Conf. on computer vision and pattern recognition}, pages 248--255. Ieee, 2009.

\bibitem{GONZALEZ-PR}
S.~Gonzalez and J.~E. Tapia.
\newblock Forged presentation attack detection for {ID} cards on remote verification systems.
\newblock {\em Pattern Recognition}, 162:111352, 2025.

\bibitem{howard2017mobilenets}
A.~G. Howard, M.~Zhu, B.~Chen, D.~Kalenichenko, W.~Wang, T.~Weyand, M.~Andreetto, and H.~Adam.
\newblock Mobilenets: Efficient convolutional neural networks for mobile vision applications, 2017.

\bibitem{poisson}
D.~H. Lee, S.~B. Yoo, M.~Choi, J.~B. Ra, and J.~Kim.
\newblock Block poisson method and its application to large scale image editing.
\newblock In {\em 19th IEEE International Conference on Image Processing}, pages 2121--2124, 2012.

\bibitem{signature-dataset-3}
M.~I. Malik, S.~Ahmed, A.~Marcelli, U.~Pal, M.~Blumenstein, L.~Alewijns, and M.~Liwicki.
\newblock {ICDAR2015} competition on signature verification and writer identification for on- and off-line skilled forgeries.
\newblock In {\em 13th International Conference on Document Analysis and Recognition (ICDAR)}, pages 1186--1190, 2015.

\bibitem{signature-dataset-2}
M.~I. Malik, M.~Liwicki, L.~Alewijnse, W.~Ohyama, M.~Blumenstein, and B.~Found.
\newblock {ICDAR} competitions on signature verification and writer identification for on and offline skilled forgeries.
\newblock In {\em 12th International Conference on Document Analysis and Recognition}, pages 1477--1483, 2013.

\bibitem{markham2023openset}
R.~P. Markham, J.~M.~E. López, M.~Nieto-Hidalgo, and J.~E. Tapia.
\newblock Open-set: {ID} card presentation attack detection using neural style transfer.
\newblock {\em IEEE Access}, 12:68573--68585, 2024.

\bibitem{dinov2}
M.~Oquab, T.~Darcet, T.~Moutakanni, H.~V. Vo, M.~Szafraniec, et~al.
\newblock {DINO}v2: Learning robust visual features without supervision.
\newblock {\em Transactions on Machine Learning Research}, 2024.
\newblock Featured Certification.

\bibitem{face-dataset-3}
N.~Strohminger, K.~Gray, V.~Chituc, J.~Heffner, C.~Schein, and T.~Heagins.
\newblock The {MR2}: A multi-racial, mega-resolution database of facial stimuli.
\newblock {\em Behavior research methods}, 48, 08 2015.

\bibitem{efficientnetv2}
M.~Tan and Q.~Le.
\newblock Efficientnetv2: Smaller models and faster training.
\newblock In M.~Meila and T.~Zhang, editors, {\em Proceedings of the 38th International Conference on Machine Learning}, volume 139 of {\em Proceedings of Machine Learning Research}, pages 10096--10106. PMLR, 2021.

\bibitem{challenge-past}
J.~E. Tapia, N.~Damer, C.~Busch, J.~M. Espin, J.~Barrachina, A.~S. Rocamora, K.~Ocvirk, L.~Alessio, B.~Batagelj, S.~Patwardhan, R.~Ramachandra, R.~Mudgalgundurao, K.~Raja, D.~Schulz, and C.~Aravena.
\newblock First competition on presentation attack detection on {ID Card}.
\newblock In {\em 2024 IEEE International Joint Conference on Biometrics (IJCB)}, pages 1--10, 2024.

\bibitem{yolo}
R.~Varghese and S.~M.
\newblock {YOLOv8}: A novel object detection algorithm with enhanced performance and robustness.
\newblock In {\em International Conference on Advances in Data Engineering and Intelligent Computing Systems (ADICS)}, pages 1--6, 2024.

\bibitem{Lora}
B.~P. Veasey and A.~A. Amini.
\newblock Low-rank adaptation of pre-trained large vision models for improved lung nodule malignancy classification.
\newblock {\em IEEE Open Journal of Engineering in Medicine and Biology}, pages 1--9, 2025.

\bibitem{face-dataset-2}
T.~Vieira, A.~Bottino, A.~Laurentini, and M.~Simone.
\newblock Detecting siblings in image pairs.
\newblock {\em The Visual Computer}, 30:1--13, 12 2013.

\bibitem{EvalAI}
D.~Yadav, R.~Jain, H.~Agrawal, P.~Chattopadhyay, T.~Singh, A.~Jain, S.~B. Singh, S.~Lee, and D.~Batra.
\newblock {EvalAI}: Towards better evaluation systems for {AI} agents.
\newblock {\em ArXiv}, arXiv:1902.03570, 2019.

\end{thebibliography}
}

\end{document}